\documentclass[10pt,twocolumn,letterpaper]{article}

\usepackage{cvpr}
\usepackage{times}
\usepackage{epsfig}
\usepackage{graphicx}
\usepackage{amsmath}
\usepackage{amssymb}
\usepackage{textcomp}
\usepackage{tikz-cd}
\usepackage{comment}
\usepackage{caption}
\usepackage{subcaption}
\usepackage{placeins}
\usepackage{authblk}

\usepackage[pagebackref=true,breaklinks=true,letterpaper=true,colorlinks,bookmarks=false]{hyperref}
\usepackage[subtle,title=normal,wordspacing = normal]{savetrees}

 \cvprfinalcopy 

\def\cvprPaperID{3159} 

\DeclareMathOperator*{\argmin}{arg\,min}

\ifcvprfinal\fi
\begin{document}

\title{\ Hard Mixtures of Experts for Large Scale Weakly Supervised Vision}

\author{Sam Gross}
\author{Marc'Aurelio Ranzato}
\author{Arthur Szlam}
\affil{Facebook AI Research (FAIR)}

\maketitle


\begin{abstract}
	Training convolutional networks (CNN's) that fit on a single GPU with minibatch stochastic gradient descent has become effective in practice.   However, there is still no effective method for training large CNN's that do not fit in the memory of a few GPU cards, or for parallelizing CNN training.  In this work we show that a simple hard mixture of experts model  can be efficiently trained to good effect on large scale hashtag (multilabel) prediction tasks.  Mixture of experts models are not new \cite{MOE,collobert_mixtures}, but in the past, researchers have had to devise sophisticated methods to deal with data fragmentation.  We show empirically that modern weakly supervised data sets are large enough to support naive partitioning schemes where each data point is assigned to a single expert.  Because the experts are independent, training them in parallel is easy, and evaluation is cheap for the size of the model.   Furthermore, we show that we can use a single decoding layer for all the experts, allowing a unified feature embedding space.   We demonstrate that it is feasible (and in fact relatively painless) to train far larger models than could be practically trained with standard CNN architectures, and that the extra capacity can be well used on current datasets.

\end{abstract}

\section{Introduction}
Large annotated image datasets have revolutionized computer vision.  The rise of data hungry machine learning methods like convolutional neural networks (CNN's) has been facilitated by training sets with millions of labeled images, especially Imagenet \cite{imagenet}.   These machine learning methods \cite{alexnet,resnets} have proven not only successful at solving the training tasks, but also for finding good features for many image tasks; and to a large extent, models that perform better on tasks like the Imagenet recognition challenge give features that are better for other tasks \cite{astounding_baseline,resnets}.

However, hand annotation is laborious.  The Imagenet dataset is small compared to the hundreds of millions of images posted to the web and social media every day.  Recent works  \cite{webly_sup, webly_cnn, weakly_supervised_vf} have shown that it is possible to build vision models with weakly supervised instead of hand-annotated data, and open the possibility of using truly gigantic datasets.

As the data gets bigger, we can expect to be able to scale up our models as well, and get better features; more data means more refined models with less overfitting.   However, even today's state of the art convolutional models cannot keep up with the size of today's weakly supervised data.  With our current optimization technology and hardware, more images are posted to photo sharing sites in a day than can be passed through the training pipeline of standard state of the art convolutional architectures.   Furthermore,  there is evidence \cite{weakly_supervised_vf, distilling} and below in this work, that these architectures are already underfitting on datasets at the scale of hundreds of millions of images.

A well established approach for scaling models in a straightforward way is to use ``mixture'' architectures, where one model acts as a ``gater'', routing data points to ``expert'' classifiers to update a final decision, for example as in  \cite{MOE,collobert_mixtures,distilling,HD-CNN,self_informed,network_of_experts}.   In this work, we make two contributions.
First, we propose a particularly simple mixture architecture where each expert is associated with a cluster in the feature space of a trained CNN, which acts as the gater.   We also describe a variant where all experts share the same decoder, allowing the feature space of the experts to be meaningful for transfer tasks.  Second, we give evidence that in the setting of weakly supervised tag prediction from images, on the large datasets that are available today,  standard CNN models are underfitting.  On the other hand, we show that despite our approach's simplicity, in this setting, it can give significant benefits to test accuracy by allowing the efficient training of much more powerful models.

\section{Models}
Denote by $x_1,..., x_N$ a labeled training set of images with target outputs  $y_1,..., y_n$. 
The basic idea of a mixture of experts \cite{MOE} model is to have a set of expert classifiers $H_1, ... , H_K$, and a gating classifier $T$.  
To evaluate the model, an input $x$ is processed by $T$, outputting a probability vector $T(x)$ with $K$ coordinates.  The output of the model is then 
\begin{equation}
	\sum_{i=1}^K T(x)_i H_i(x) 	
\end{equation}
Here, $T$ and $H_i$ will be convolutional networks as in \cite{alexnet,resnets}.  Moreover, we will consider the simple situation where for each $x$,  $T(x)$ is nonzero in just one coordinate.   Because our models will pick a single expert instead of a distribution over the experts, our models are ``hard'' mixtures.

In this work,  we will  not train our models end to end.  That is, we will not attempt to directly optimize $T$ so that the assignment of $x$ to an expert minimizes the final classification loss.  Instead, we will build  $T$ as follows:
first, train a standard supervised convolutional network $\tilde{T}$ with $L$ layers to produce $y_j$ from $x_j$.  Once we are satisfied with the optimization of $\tilde{T}$, construct $z_j = \tilde{T}^{L-1}(x_j)$, that is, take $z_j$ to be the output of the last hidden layer of $\tilde{T}$  before the decoder for each training point $x_j$.  We then do $K$-means clustering on the $z_j$, obtaining cluster centers $c_1,..c_K$.  Then 
\begin{equation}
T(x)_i =
\begin{cases}
    1,  & \text{if } i = \argmin_j ||\tilde{T}^{L-1}(x)-c_j||_2 \\
    0,  & \text{otherwise}
\end{cases}
\end{equation}  
Our model is thus a ``local'' architecture, in the sense that a location in the feature space defined by $\tilde{T}^{L-1}$ corresponds to a choice of expert classifier.  Note that we make no attempt to balance the number of images per  each cluster.

\begin{figure}
\centering
\begin{tikzcd}[ampersand replacement=\&]
 \&  \& Y   \&   \&   \\
 \&  \& D \arrow{u}  \&   \&   \\
H_1 \arrow{urr} \& H_2 \arrow{ur} \& H_3 \arrow{u}  \&  H_4 \arrow{ul} \& H_5\arrow{ull}  \\
     \&  \& T \arrow{ull} \arrow{ul} \arrow{u} \arrow{ur}  \arrow{urr}  \&  \&   \\
 \& \& X \arrow{u} \& \&
\end{tikzcd}
\caption{A mixture of experts model with a shared decoder.  Here $Y$ is the labels, $H_i$ are the experts, and $D$ is the decoder.  Since the decoder is shared, the images are mapped into a shared embedding space.}
\label{fig:sharedD}
\end{figure}
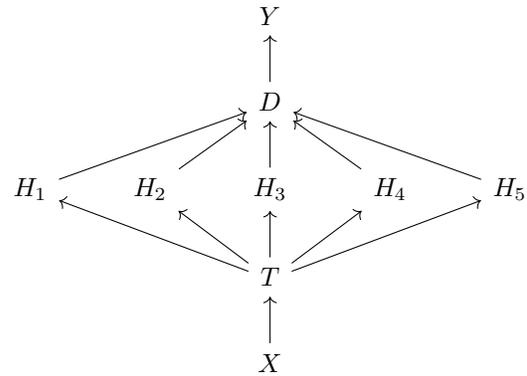

Once we have $T$, which given an input $x$ outputs an expert $i$, we have two possible methods for building the rest of the model.  In the simplest version, each $H_i$ outputs a distribution over the $y$, leading to a model as in Figure \ref{fig:independentD}.  In this case, we can optimize the $H_i$ independently, each on its own share of the training data.   This model is useful if at test time, the only thing we care about is predicting the labels $y$.  However, it is often the case that we are training the model because we want the features in the last layer before the decoder, rather than label predictions.  In this case, we keep each $H_i$ independent, but instead of outputting probabilities over labels, each $H_i$ outputs a feature vector, and we append a shared decoder $D$ to the model.  See Figure \ref{fig:sharedD}.  Note that in both of our models, the output for each expert is a distribution over the full set of possible labels.

Training a model with a shared decoder is more involved than the model with independent decoders, but if there are a large number of classes, the gradients from $Y$ to $D$ are relatively sparse.  In this setting, we use one machine to hold $D$, and one machine for each $H_i$.

\begin{figure}
\centering
\begin{tikzcd}
Y_1 & Y_2 & Y_3  &  Y_4 & Y_5  \\
H_1 \arrow{u} & H_2 \arrow{u} & H_3 \arrow{u}  &  H_4 \arrow{u} & H_5\arrow{u}  \\
     &  & T \arrow{ull} \arrow{ul} \arrow{u} \arrow{ur}  \arrow{urr}  &  &   \\
 & & X \arrow{u} & &
\end{tikzcd}
\caption{A mixture of experts model with separate decoders for each expert.  Here $Y_i$ is the labels associated to points assigned to the $i$th expert, and  $H_i$ are the experts.}
\label{fig:independentD}
\end{figure}
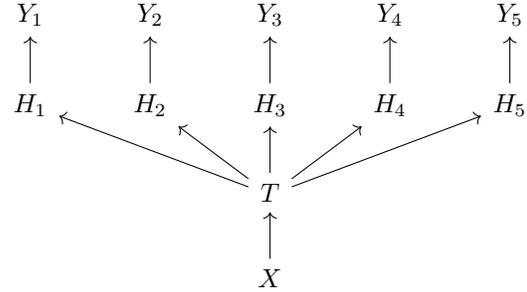

\subsection{Advantages and liabilities of hard mixtures of experts versus standard CNN's}
The models proposed above have some important scalability advantages compared to a standard CNN with $K$ times as many feature maps.  They are efficient in terms of wall clock time per parameter at train and test:  because each $H$ is trained independently, it is easier to parallelize.  Furthermore, independent of the number of $H$, at evaluation time, the cost of finding an output from an input $x$ is the cost of computing $T(x)$ plus the cost of computing $H_i(x)$ for a single $i$; whereas for a large CNN with $K$ times as many feature maps, a naive forward could cost as much as $K$ times as much.  Although many methods exist for compressing the layers to make evaluation faster (and have smaller memory footprint), there has been less success training models in compressed form, and it would be difficult with current technology to train CNN's as large as the ones we discuss here.

On the other hand, the models described here are inefficient compared to a standard CNN in terms of modeling power per parameter and in terms of data usage.  Because each $H_i$ acts independently of the others, the parameters in $H_i$ do not interact with the parameters in $H_j$.   This is in contrast with a standard CNN, where each parameter can interact with any other.  Moreover, because the training data is split amongst the experts, each parameter sees only a fraction of the data as it is trained (although in the model with the shared decoder, the decoder parameters see all the data).

While there are many problem settings where data efficiency is key, it is also the case that there are problems where data is cheap, and we want to train the largest possible model to most accurately fit the data.  In these situations, one quickly runs up against hardware and algorithmic limitations of training a standard CNN with serial stochastic gradient descent, and hard mixtures of experts become attractive.    In particular, to make a CNN that has $K$ times as many parameters, it is infeasible to scale the number of feature maps by $K$, but making a $K$ expert hard mixture model is practical.

\begin{table*}
\centering
\begin{tabular}{l |c  c  c  c  c | c  c  c  }
Model                & Train Loss & Test Loss    & q@1     & q@5     & q@10       & p@1   &  p@5   & p@10 \\
\hline
ResNet-18   	      &  7.78       &  7.78     &  3.04\% & 8.69 \% &  12.41\%	& 1.38\% &  4.94\% & 8.01\% \\
ResNet-34   	      &  7.71     &  7.72     &  3.31\% & 9.59 \% &  13.80\%	& 1.47\% &  5.31\% & 8.62\% \\
ResNet-50   	      &  7.65     &  7.66     &  3.47\% & 9.80 \% &  13.88\%	& 1.55\% &  5.49\% & 8.86\% \\
ResNet-50 $4\times$feature size
                     &  7.65     &  7.70     &  3.80\% & 10.49 \% &  14.74\%	& 1.71\% &  5.96\% & 9.52\% \\                     
                     
ResNet-18 ensemble-50 &  7.62      &  7.66     & 3.37\% & 9.43\% & 13.38\%	& 1.56\% &  5.53\% & 8.90\% \\
\hline
ResNet-18 MoE-25	  &  7.03      &  7.10     & 5.35\% & 14.53\% & 19.76\%	& 2.21\% &  7.64\% & 12.00\% \\
ResNet-18 MoE-50	  &  7.03       &  6.93     & 6.12\% & 16.27\% & 21.74\%	& 2.48\% &  8.64\% & 13.48\% \\
ResNet-18 MoE-75	  &  6.72       &  6.84     & 6.65\% & 17.40\% & 23.33\%	& 2.62\% &  9.15\% & 14.26\% \\
ResNet-18 MoE-100	 &   6.32    &  6.81     & 6.87\% & 17.88\% & 23.82\%	& 2.69\% &  9.47\% & 14.75\% \\
ResNet-34 MoE-50	 &   6.49    &  6.78     & 6.77\% & 17.72\% & 23.76\%	& 2.70\% &  9.43\% & 14.70\% \\
ResNet-18 MoE-50 shared decoder
                    &   6.97     &  7.13     & 5.67\% & 14.60\% & 19.70\%	& 2.24\% &  7.89\% & 12.35\% \\
\hline
ResNet-34 MoE-50 oracle	 &   5.57    &  5.65     & 9.7 \% & 23.8\% &  32.8\%	& NA &   NA & NA \\
              
\end{tabular}
\caption{YFCC100M hash tag prediction results. p@m and q@m are computed as in Equations \eqref{eq:p} and \eqref{eq:q} respectively. ``ResNet-18'' refers to a ResNet with 18 layers, and MoE-$a$ refers to a model with $a$ experts.  If the base model has $P$ parameters, the  MoE model with $a$ experts has $P(a+1)$ parameters.  The $4\times$feature size model has hiden layers 4 times as many hidden units as a normal ResNet.   The ResNet-18 MoE-50 shared decoder model has about 36 times as many parameters as its base model.  Both cost twice as much to evaluate as their base models (as the expert is the same size as the base model) and perform significantly better.   Examples of predictions are shown in Figure \ref{fig:qualitative_tags}.   The ``oracle'' model  uses the best possible choice of expert using the true test label; we do not compute it for p@m because the notion of ``best'' is more involved.} 
\label{tab:flickr_tag}
\end{table*}

\subsection{Advantages and liabilities of no end-to-end training}
End-to-end training may lead to more accurate models, and at test time, such systems could be made as efficient as the ones presented here. 

However, at the scale discussed in this paper, end-to-end training of mixture models continues to be an exceptional engineering endeavor, and the computing infrastructure necessary to make it work is not yet widespread (or even well developed).  On the other hand, the techniques described in this work are simple, and can be used by any lab with a GPU, as each expert trains independently.  Because the experts train much faster than the trunk, the total time, even trained in serial, is a small multiple of training the trunk. 

Concurrently with this work, \cite{OLNN} described a distributed end-to-end mixture of experts system for training RNN's in NLP settings. This shows that it {\it is} possible to train mixture models end-to-end at scale; here we show that it can be valuable to take a simpler approach.




\section{Related Work}
This paper is built on two columns of ideas: weakly supervised training of convolutional networks with very large datasets, and scaling machine learning models via mixtures of experts.

 Recent works have shown that noisy tags and captions from large image collections can be an effective source of supervision, e.g. \cite{webly_sup, webly_cnn, weakly_supervised_vf} and the references therein.  These works pave the way for using web-scale training sets for learning image features.  In this work, we use the same framework as in these, but we show that we can improve results with larger capacity models.

The model we propose is a particularly simple mixture of experts.  These models were introduced in \cite{MOE}.  Our differs from this and many following works in that we use a hard mixture for efficiency and scale, rather than a soft probabilistic assignment.   This is very similar to the approach in \cite{collobert_mixtures}.  However, in this work, instead of using multiple rounds of optimizing the gating with the classifiers fixed, and then optimizing the classifiers, we use a single round of $K$-means on the feature outputs of a non MoE classifier to get the  expert assignments.  Furthermore, both the gater and the experts are convolutional networks, instead of SVM's.  

Recently there have been several works using mixture of experts type models with CNN's for vision tasks  \cite{distilling,network_of_experts,self_informed,HD-CNN}.   Roughly, these models and ours differ in how they deal with routing data to experts, which parts of the model are shared, and how results from the experts are combined.  Our work is similar to these in that it uses a ``generalist'' convolutional network trained for predicting labels as a gater.

 In each of these works, data is routed to the experts by an agglomeration of the classes into abstract superclasses.  The gater sends an image to an expert based on which superclass the gater thinks the image is in.   In our work, the output of the final hidden states of the gater applied to the training data are clustered directly, as opposed to the class labels.  This is more suitable for the tag prediction tasks we focus on here, as an image can have multiple different tags.  while it is certainly true that certain tags co-occur more than others, it is still the case that one would like to send an image to an expert based on the set of likely tags, rather than any particular tag; and because we focus on scaling the model, we do not want to have to look at all experts appearing in the union of the possible tags for an image.   Even in the single-label image categorization setting, clustering the image embedding rather than the classes may partially mitigate errors where the gater makes an unrecoverable mistake.  We have found that we can get good results with each image being mapped to a single expert.  
 
In \cite{distilling,network_of_experts,self_informed,HD-CNN} different methods of combining the output from the experts are proposed.  In \cite{distilling}, the authors propose solving an optimization at evaluation time to match the output distributions of the generalist and the experts.  In \cite{self_informed}, the expert networks are taken as parallel layers and mapped directly onto the outputs with the generalist.  In \cite{network_of_experts,HD-CNN}, the output of the model is the distribution given by taking the weighted some of the distributions of the experts by the distribution given by the generalist over the coarse classes.   Our model as in Figure \ref{fig:independentD} can be considered as a simple form of \cite{network_of_experts,HD-CNN}, in the sense that the generalist outputs a delta, and we simply take the output of the relevant expert as the output.  Our model as in \ref{fig:sharedD} is somewhat different than any of these, as an expert outputs {\it feature vector} that is fed to a shared decoder.
 

Concurrent with this work is \cite{OLNN}, which also puts a hard mixture model as a component in a deep learning architecture for large scale language modeling.  In this work, we work on a different problem domain, and we use a simpler gating scheme.


\section{Experiments}
\begin{table*}[h!]
\centering
\begin{tabular}{l | c  c  c  c  c  c  c }
  Model & Actions & Flowers & Birds & MIT SUN & Indoors & Sports  & Imagenet\\
   \hline
   AlexNet       & 51.69 & 69.72 & 22.69     & 42.67 & 53.19 & \textbf{91.3} & 34.3\\
  ResNet-18     & \textbf{53.15} & 64.76 & 21.28 & 45.01 & 55.38 & 84.4 &  36.9\\
\hline
  ResNet-18 MoE & 49.69 & \textbf{76.93} & \textbf{33.71} & \textbf{45.15} & \textbf{57.89} & 78.3 & \textbf{42.1} \\
\end{tabular}
\caption{Flickr transfer results; the numbers are the test accuracies on each of the datasets except for Imagenet, where the reported number is test accuracy on the validation set.  The MoE model has 50 experts.  All models were trained on YFCC100M, then the decoder layer was removed (and the rest of the model fixed), and then the decoder was retrained on the training labels for that dataset.  The ResNet-18 and ResNet-18 MoE models have 512 dimensional feature representations, and the AlexNet has a 4096 dimensional feature representation.}
\label{tab:transfer}
\end{table*}
\begin{table*}[t]
\centering
\begin{tabular}{c}
\includegraphics[width = .95\linewidth]{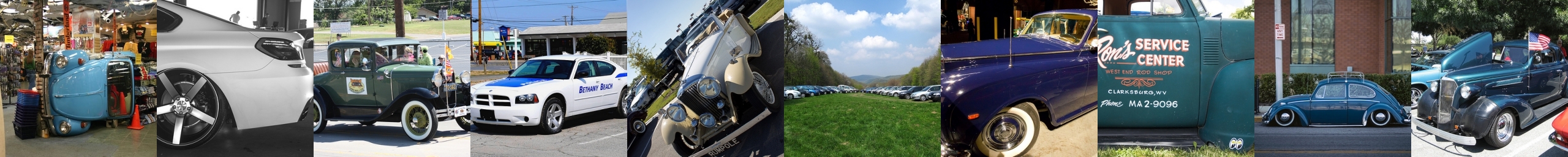} \\
\includegraphics[width = .95\linewidth]{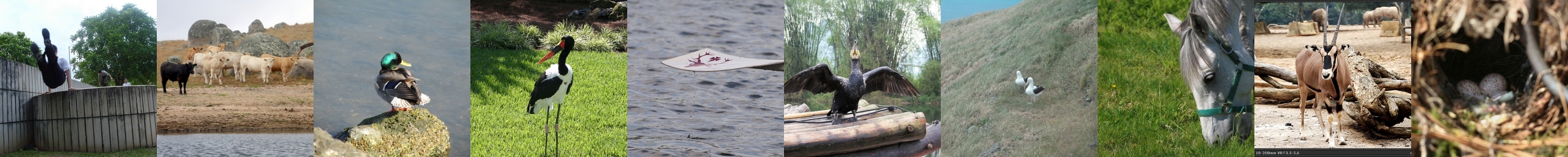} \\
\includegraphics[width = .95\linewidth]{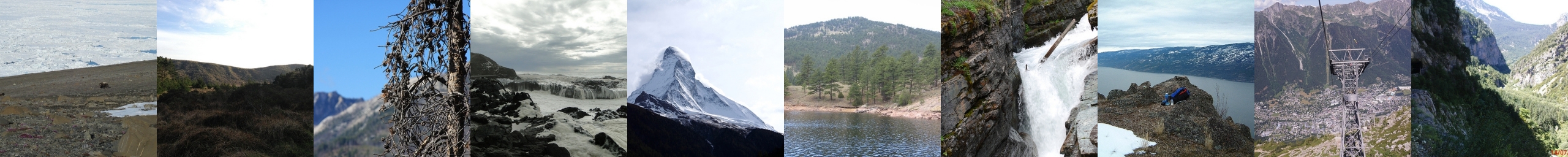} \\
\includegraphics[width = .95\linewidth]{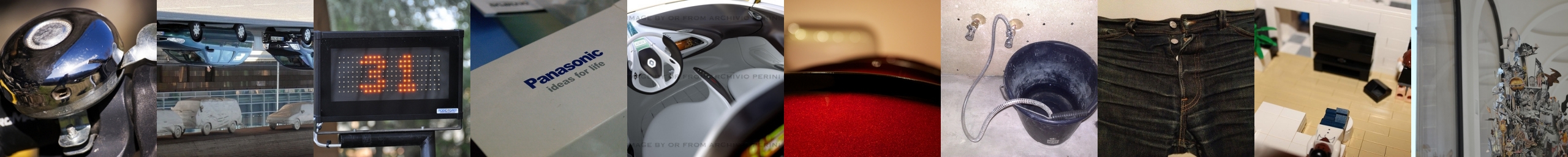} \\
\end{tabular}
\caption{Visualization of 50 clusters from the feature space of a ResNet-18 model.  Each row corresponds to a cluster; the images in that row a randomly sampled from the cluster.
The first row is cluster 50, which is a peak for the tag ``porsche'' in Table \ref{tab:tag_cluster_hist}.  This cluster has ~1.3\% of the images in the train set.  The next rows are, respectively: cluster 16, which is a peak for the tag ``zebra''  and has ~2.2\% of the images in the train set;  cluster 26, which is a peak for the tag ``park'' and has ~3.4\% of the images in the train set;  
cluster 3, which is a peak for the tag ``keyboard'' and has ~1.8\% of the images in the train set. The distribution of the sizes of the clusters is shown in Figure \ref{fig:cluster_util_hist}. }
\label{tab:cluster_reps}
\end{table*}

We will discuss three sets of experiments.  In the first, we train and test on 1000 category labeled Imagenet \cite{imagenet}.   In the second set of experiments, our models will be trained on (weakly supervised) tag prediction following the training procedures in \cite{weakly_supervised_vf}, and we report the tag prediction results.   In the final set of experiments, the models are trained for tag prediction, and all of the model except the last layer is fixed.  The last layer is then trained on new labeled datasets to see how well the weakly supervised features can be used for transfer learning.  We use the following datasets for tag prediction:

\noindent{\bf YFCC100M:}
The YFCC100M \cite{flickr100m} contains roughly 100 million color images collected from \url{https://www.flickr.com/} with captions and hashtags.  We build a dictionary of $M= 10000$ words by sorting all tokens appearing in the dataset by frequency, and keeping those with rank greater than 500 and less than 10500.

\noindent{\bf Instagram Food:}
We also collected a set of 440 million anonymized images from instagram (\url{https://www.instagram.com/}) that contain words relating to food in their hashtags or captions.   The dictionary of words used to select the images was obtained by starting with a few seed words (e.g. ``breakfast'', ``lunch'', ``dinner'', ``yum'', ``food''), and finding images that contained these words in their captions or hashtags.  Then the co-occurring words were kept based on tf-idf score (here a ``document'' is the caption and tags associated with an image).  

We are not able to release this data set, and so the community will not be able to reproduce the results we report.  Nevertheless,  we think the results are valuable as anecdotal evidence.

\subsection{Weakly Supervised Training Loss}
As in \cite{weakly_supervised_vf}, we take $y_i$ to be a $M$ vector with a $1$ in the $j$th entry if $x_i$ contains the tag $j$, and we use the loss function 
\begin{equation}
L(y,\hat{y}) = -\sum_{j = 1, M} y_{ij}\log\left(\frac{\exp(\hat{y}_{ij})}{\sum_{j'}\exp(\hat{y}_{ij'})} \right)
\end{equation}
 where $\hat{y_i} = H_{T(x)}(x)$ if we are using the independent heads as in Figure \ref{fig:independentD} or $\hat{y_i} = DH_{T(x)}(x)$ if we are using the shared decoder as in \ref{fig:sharedD}

\subsection{Model and training details}
For each model we train, we use the same architecture for the  gater $T$ as for the experts $H_i$.   We choose between Alexnet \cite{alexnet} and ResNet\cite{resnets}; so if $T$ is a resnet, each $H_i$ is as well.  On each dataset, we first train $T$ until the error plateaus.  We use $K = 50 $ centroids in the $K$-means computation.  Before running the clustering on $Z = T^{L-1}(X)$, we project it to its first $256$ PCA dimensions.   

We train $T$ using stochastic gradient descent with minibatch size 256 with weight decay 0.0001 and momentum of 0.9. For training, we define an "epoch" as one million images. We start the learning rate at 0.1 and divide it by 10 every 60 epochs. We train each $H$ in the same manner, but without momentum and we divide the learning rate by 2 every 5 epochs.

We follow \cite{weakly_supervised_vf} and sample {\it per class} at training.  That is, we pick a word in the dictionary at random, then pick an image having that word as a tag.  All other tags for that image are counted as negatives for that example.

Corroborating the reports of \cite{distilling}, we have found that the experts train faster than the original model.   On YFCC100M, the gater $T$ takes 3 to 5 days to train on 8 GPUs; each $H$ takes less than $16$ hours on 1 GPU.  Thus if the experts are trained in parallel, the total training time is less than 6 days.  Roughly, we train the generalist for 200 epochs and each expert for around 20 (the exact stopping time is determined by the validation loss).  This means that even without a large number of GPUs it is possible to train the mixture model for tens of experts without a small multiple of the total training time.


\subsection{Imagenet results}
Imagenet \cite{imagenet} is a hand annotated dataset with 1.2 million color images, roughly evenly taken from 1000 categories.  We include Imagenet experiments to show that our simple architectures do {\it not} improve accuracies at this scale (as opposed to the scale of the YFCC100M dataset); note however that by combining the results of the experts with more care, the authors of \cite{HD-CNN,network_of_experts} were able to improve results on Imagenet.  Our results are in Table \ref{tab:imagenet}.

\begin{table}
\centering
\begin{tabular}{l | c  c }
   Model & Top-1 error & Top-5 error  \\
\hline  
  ResNet-18 & 30.64 & 10.69\% \\
   ResNet-18 MoE-50 & 30.43 & 11.7\%
\end{tabular}
\caption{Imagenet classification results.  The mixture of experts model uses 50 experts.  Error results are reported on 50\% of the validation data (the other 50\% was used to determine hyperparameters).  We do not see improvements in accuracy at this scale.}
\label{tab:imagenet}
\end{table}

\subsection{Tag prediction results}

\begin{figure}
\captionsetup[subfigure]{labelformat=empty}

 \centering
  \subcaptionbox{{\bf True tags}: {\tiny book beauty view hat hibiscus nature blues take look bernard white turns pink grown flat yellow flowers indoor warm reds fill flower pretty plants shaw gear great colours} \newline {\bf Predicted}: { \scriptsize hibiscus rosa petals flor pink naturesfinest fleur blume blooms flower}}{\includegraphics[width=.2\textwidth]{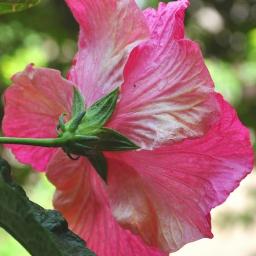}}\hspace{1em}%
  \subcaptionbox{{\bf True tags}: {\small poca \bigskip} \newline {\bf Predicted}: { \scriptsize earlybird instagr square bookstore flicksquare instagood likes sq cafa espresso}}{\includegraphics[width=.2\textwidth]{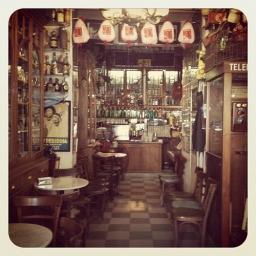}}

    \subcaptionbox{{\bf  True tags}: {\scriptsize informatie aan zee september} \newline {\bf Predicted}: { \scriptsize keynote symposium audience speaker lecture graduate auditorium relations commencement tedx}}{\includegraphics[width=.2\textwidth]{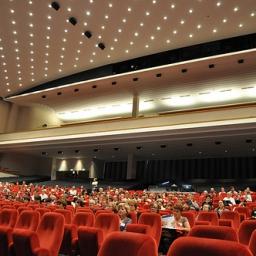}}\hspace{1em}%
  \subcaptionbox{{\bf True tags}: {\scriptsize primavera fleur fiori milano flower} \newline {\bf Predicted}: { \scriptsize plum berries blossoms cherry blossom sakura blooming branches shrub petals}}{\includegraphics[width=.2\textwidth]{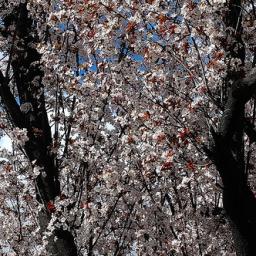}}

    \subcaptionbox{{\bf  True tags}: {\scriptsize convention las vegas nevada centre construction} \newline {\bf Predicted}: { \scriptsize construction condo crane cranes uptown reconstruction skyscraper completion apartments elevated}}{\includegraphics[width=.2\textwidth]{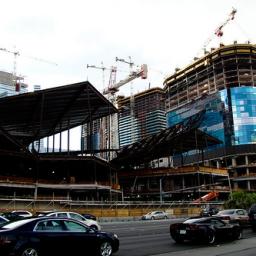}}\hspace{1em}%
  \subcaptionbox{{\bf True tags}: {\scriptsize preto brazil travelling} \newline {\bf Predicted}: { \scriptsize medellin nicaragua medella kerala amazonas parana passeio belo gerais jardim}}{\includegraphics[width=.2\textwidth]{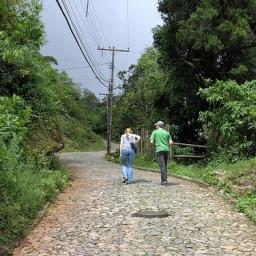}}

\caption{Some qualitative results on tag prediction on the test set of YFCC100M using ResNet-18 MoE-100 model.  Note that the true tags are often noisy.}
  \label{fig:qualitative_tags}
\end{figure}

Before displaying the results, we need to define two statistics describing model accuracy.  First, given a vector $y$, define $t_m(y)$ to be the vector with the top $m$ largest values of $y$ set to $1$ and all other entries set to $0$.  Given a set of test images $x_i$ with true targets $y_i$ and model outputs $\hat{y}_i$ (recall $y$ is the indicator of the tags that are associated with $x$), define  
\begin{equation}
\text{p@m} = \frac{\sum_i t_m(\hat{y_i})^Ty_i}{\sum_i\sum_j y_i(j)}. 
\label{eq:p}
\end{equation}
Note that the denominator is the sum of all the tags associated to all the test images (and where a tag is counted for each image it is associated with).   This measure is used in \cite{weakly_supervised_vf}, and we include it for comparison with their work.

However, this is not how our models were trained -- a model gets a disadvantage with this measure by downweighting frequent terms, as we do in our training.  Thus we also use the statistic defined as follows:  pick a number of samples $S=100000$, and for each sample pick tag $j$ uniformly from the dictionary and an image $x$ such that $j$ is a tag for $x$.  Define 
\begin{equation}
\text{q@m} = \frac{1}{S}\sum_{\text{sampled $x$, $j$}} t_m(\hat{y})(j).
\label{eq:q} 
\end{equation}
This statistic is computed in the same way we trained our models.

In table \ref{tab:flickr_tag} we show the results of the tag prediction on YFCC100M, and we show qualitative results in Figure \ref{fig:qualitative_tags}.   In addition to reporting the p@m and q@m as defined above, we also report the average test loss where the tag is sampled uniformly from the dictionary, and the image is sampled uniformly conditioned on it having that tag.

We can see that for both shared decoders and independent decoders, and for each base CNN architecture, the mixture of experts model has significantly better test loss,  better q@m, and better p@m than the base model.  This is true for both test sampling schemes, even though the model was trained with the first one.  Moreover as we increase the number of experts, from 25 to 100, the test q@m and p@m increases.  

We also compare against a pure ensemble of 50 models.  We can see that the mixture model is getting more gains from the extra parameters (and indeed, evaluating the ensemble costs 50 times as much as evaluating the base model, whereas evaluating the hard mixture of 50 experts costs 2 times the base model)

\begin{figure}
\captionsetup[subfigure]{labelformat=empty}
 \centering
 \includegraphics[width=.45\textwidth, height=.45\textwidth]{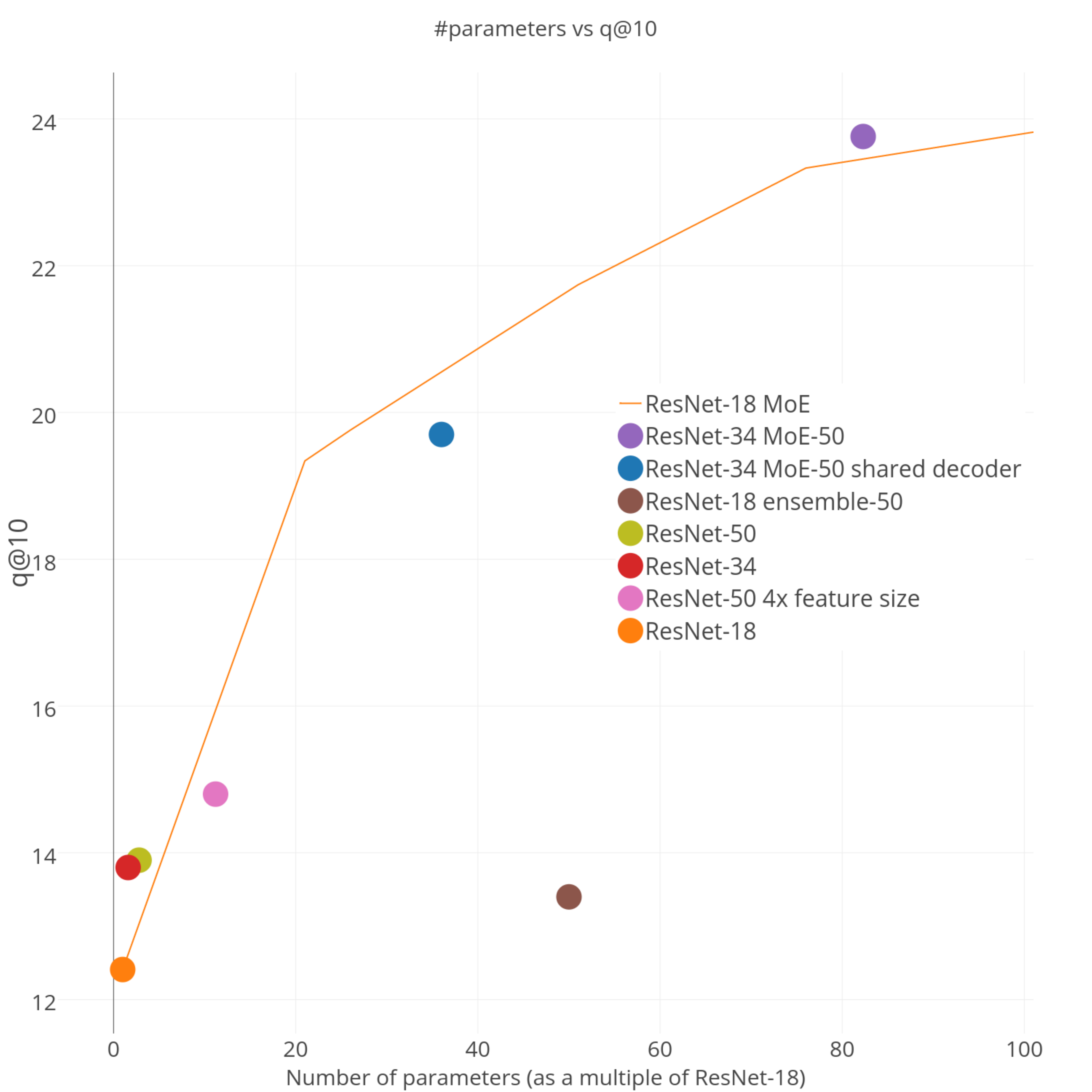}
\caption{Number of parameters vs. test accuracy for many models on YFCC100M.  
Accuracy is measured in q@10.  The orange curve corresponds to ResNet-18 MoE models.  Except for the ensemble model, which is far less accurate per parameter, all other models lie roughly on the curve, and test accuracy increases with model size.  Our approach allows training large models that can take advantage of the rich data.}
  \label{fig:params_v_q}
\end{figure}

We see evidence that the base models are underfitting: their train loss is almost the same as their test loss; and their test loss is much worse than the larger capacity mixture models.  The same trends are visible in Table \ref{tab:food_tag}.  The base models seem to be underfitting, and the larger models are able to do significantly better.    Another view on this is in Figure \ref{fig:params_v_q}.  There we plot the number of parameters of the various architectures against the test accuracies.    

The shared decoder does incur a loss of accuracy compared to independent decoders.  While it is possible that this is due to the shared decoder model being less powerful than the independent decoder model, (and Figure \ref{fig:params_v_q} supports that view), another possibility arises from the fact that in the independent decoder model, we can do early stopping on each expert individually based on its validation loss; but in the shared decoder model, it is not easy to have individual early stopping.

In the last row of table \ref{tab:flickr_tag}, we include an ``oracle'' mixture of experts model that gets to use the test tags to choose its experts.  This gives some sense of how much is lost due to not training end-to-end, as presumably end-to-end training could allow more accurate choice of experts.  We can see that there is a lot of room to more accurately choose experts.  Note that this is not an upper bound, because end-to-end training may also allow the experts to more efficiently specialize.  However, in our experience, there is a lot of information in the assignments of experts to inputs, and recovering this with an end-to-end model will not be trivial (aside from the engineering issues of training such a model at this scale).

Finally, we warn the reader that in both the YFCC100M and the Instagram food datasets, the train, test, and validation splits are just random subsets of the images.  Therefore, image-sets from a single user can be divided across the splits, which allows some over-fitting without penalty (for example, if a user takes a large number of images of a single event).  However, in our view, as datasets get larger, we will need models that have the capacity to over-fit on data of this size, which the standard models cannot do.  Moreover, there are many settings (for example, for retrieval) where this kind of ``over-fitting'' is a feature.

\begin{table*}
\centering
\begin{tabular}{l | c  c  c  c  c | c  c  c  }
Model                & Train Loss & Test Loss    & q@1     & q@5     & q@10       & p@1   &  p@5   & p@10 \\
\hline
ResNet-18   	      &  7.16       &  7.16     &  4.25\% & 10.67 \% &  15.24\%	& 5.36\% &  13.95\% & 19.91\% \\
ResNet-34   	      &  6.96       &  6.97     &  4.88\% & 12.08 \% &  16.82\%	&  5.65\%   &  14.66\% & 20.83\% \\
\hline
ResNet-18 MoE-50	  &  6.60      &  6.47     & 7.07\% & 16.06\% & 21.36\%	& 7.22\% &  18.28\% & 26.14\% \\
ResNet-34 MoE-50	  &  6.42       &  6.25     & 8.13\% & 17.89\% & 23.54\%	& 7.39\% &  18.96\% & 27.45\% \\
\end{tabular}
\caption{Instagram food hash tag prediction results. p@m and q@m are computed as in Equations \eqref{eq:p} and \eqref{eq:q} respectively. ``ResNet-18'' refers to a ResNet with 18 layers, and MoE-$a$ refers to a model with $a$ experts.  If the base model has $P$ parameters, the  MoE model with $a$ experts has $P(a+1)$ parameters and costs twice as much time to evaluate.}
\label{tab:food_tag}
\end{table*}

\subsection{Analysis of YFCC100M clusters}

In this section we fix a ResNet-18 model and a $K=50$ clustering of the features from that model.
Because the model is already achieving non-trivial tag prediction, the clusters have structure both in terms of the images and the tags.

In Table \ref{tab:tag_cluster_hist} we select a few tags, and plot the distribution of those tags over 50 clusters from the features of a ResNet-18 model.  We can see that the clusters cover different kinds of vocabulary.  Some words are mostly concentrated in very few clusters, but some words are spread over many.  Most words appear in more than one cluster.

\begin{figure}
\includegraphics[width = .95\linewidth]{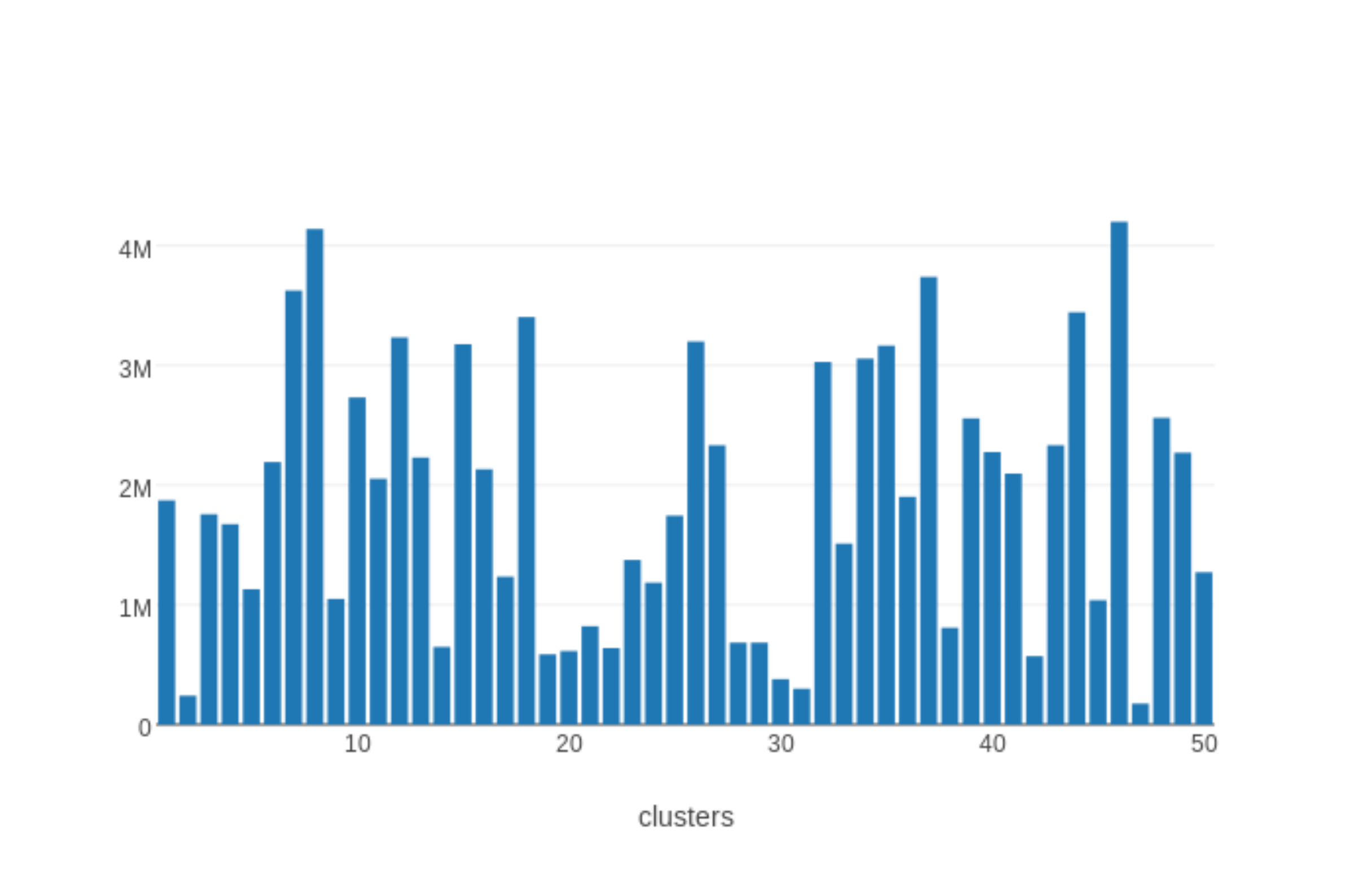}
\caption{In a 50 expert model with ResNet-18 base, the number of images routed to each cluster.  A few of the clusters are quite small, but all have non-trivial utilization}
\label{fig:cluster_util_hist}
\end{figure}

In Figure \ref{fig:cluster_util_hist} we show the distribution of the sizes of the clusters.  As mentioned above, we made no attempt to force the clusters to have equal numbers of images; thus there is a wide range of cluster sizes.  However, all the clusters are utilized.

In Table \ref{tab:cluster_reps} we display some random images from random clusters, to give a qualitative sense of the composition of the various clusters.   In the supplemental, we display random samples from all of the clusters.  

\begin{figure}
\includegraphics[width = .95\linewidth]{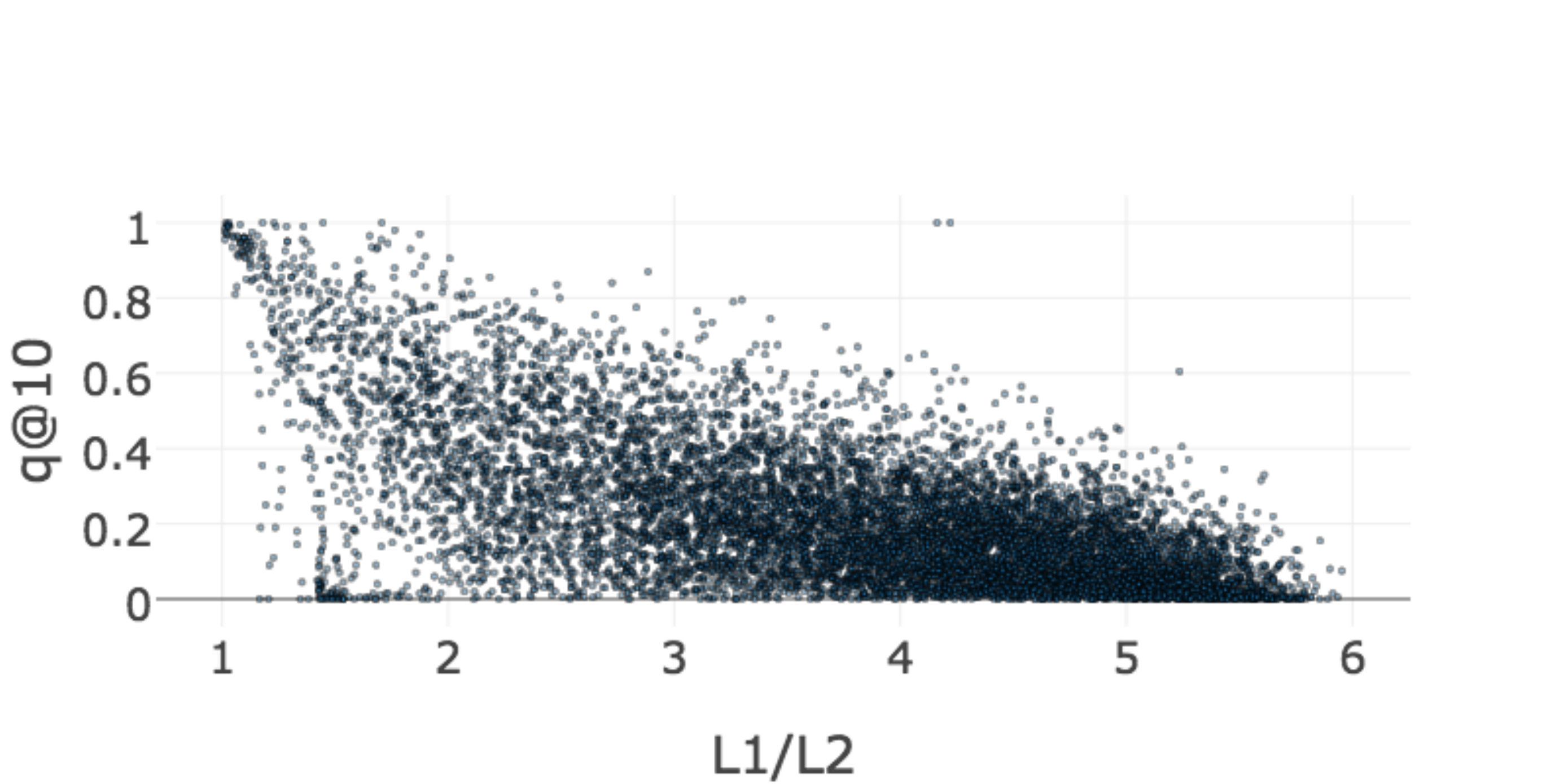}
\caption{Sparsity of cluster distribution vs. accuracy.   Each dot represents a tag $t$; let $c(t)$ be the distribution over clusters of the tag $t$.   The $x$ axis is $||c(t)||_1/||c(t)||_2$, and the $y$ axis is q@10 for that tag.   Roughly: our models are more accurate at predicting tags with sparser distributions over the clusters.}
\label{fig:cluster_sparsity}
\end{figure}

In Figure \ref{fig:cluster_sparsity}, for each tag,  we plot the sparsity of the distribution of that tag over clusters against the ResNe-18 MoE-50 accuracy for that tag as measured by q@10.  We can see that tags that have a sparse distribution over clusters are often easier for the model to infer.

\begin{table*}
\centering
\begin{tabular}{cccc}
\includegraphics[width = .23\linewidth]{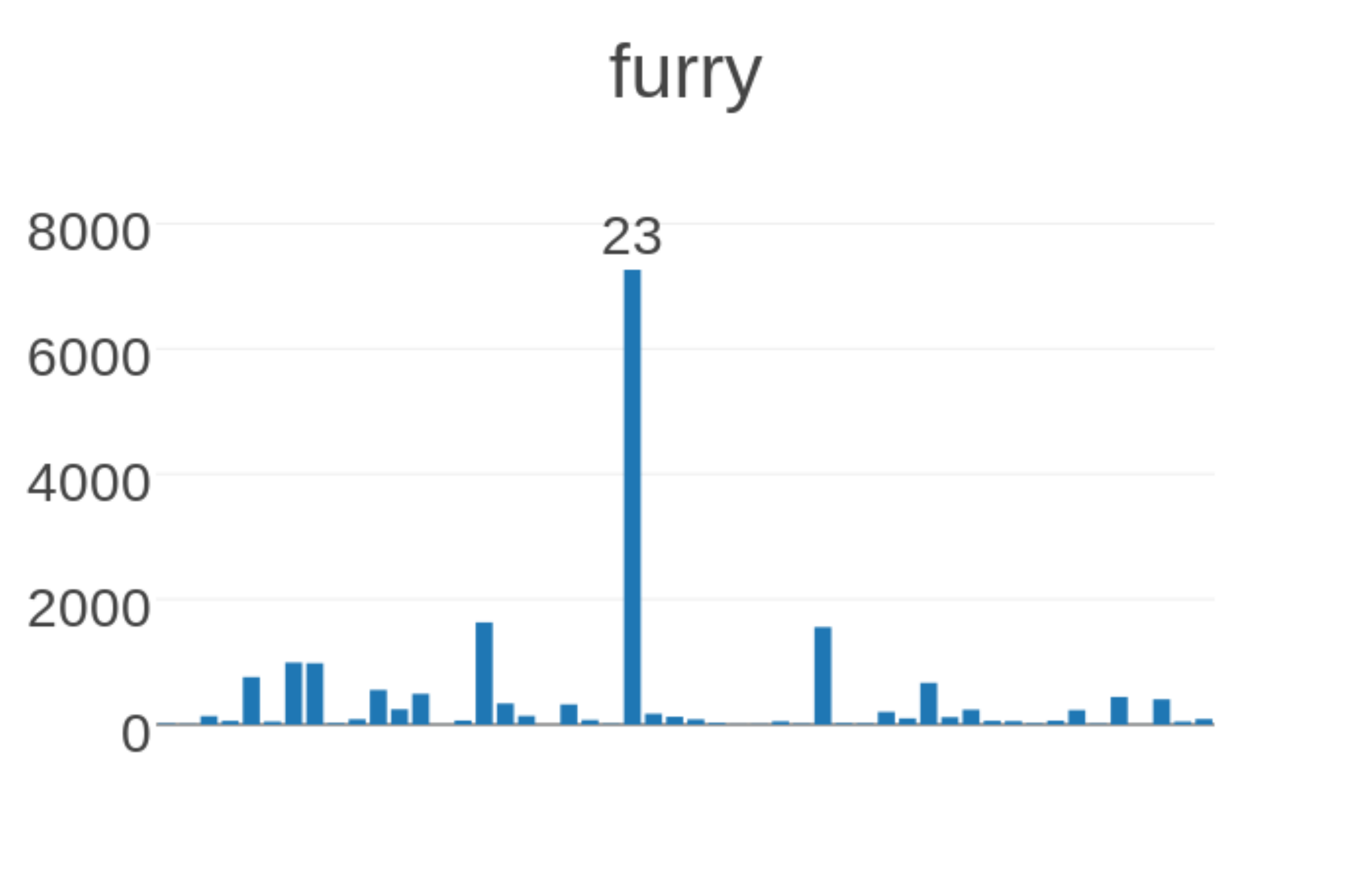} & \includegraphics[width = .23\linewidth]{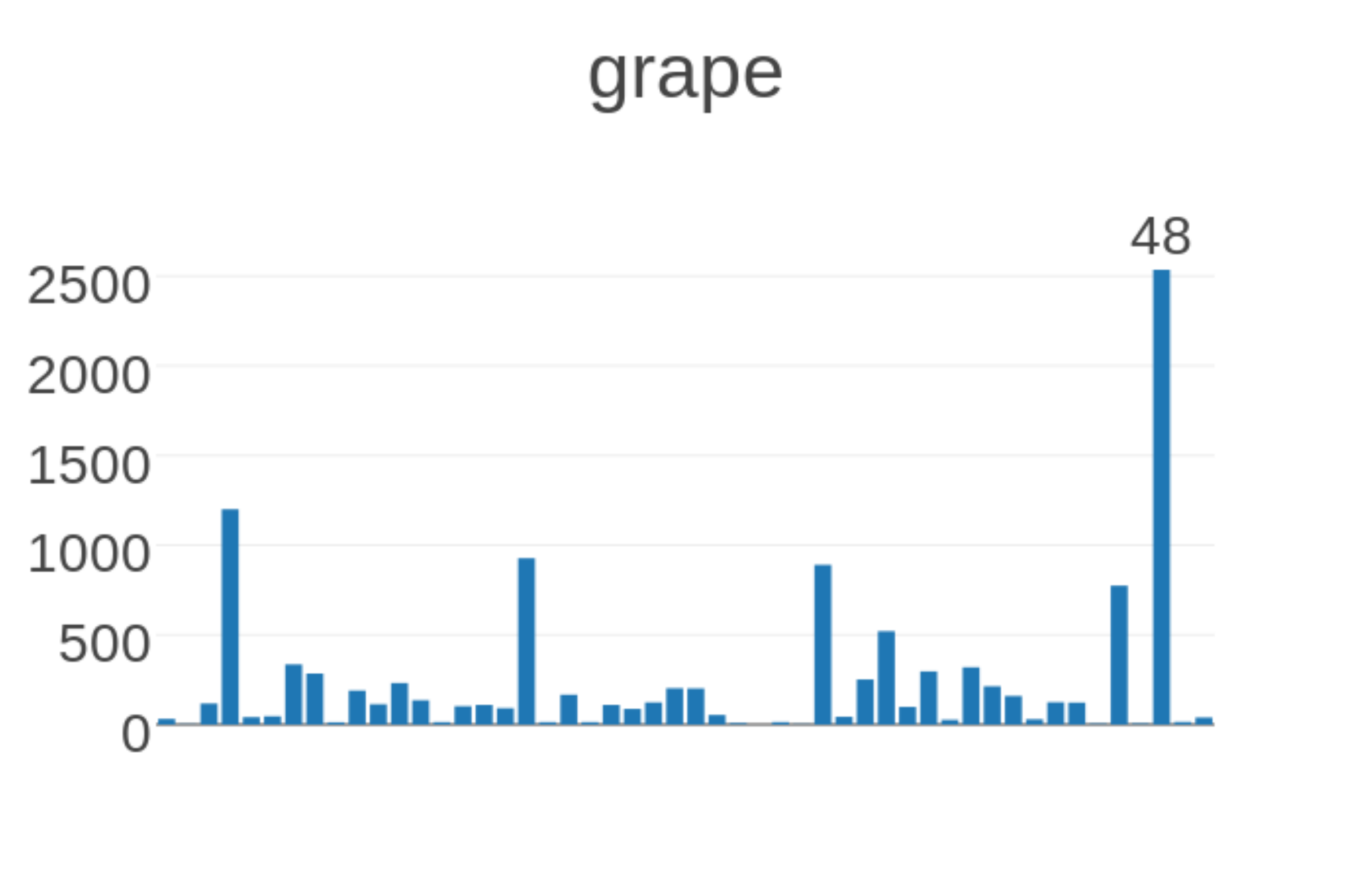} & \includegraphics[width = .23\linewidth]{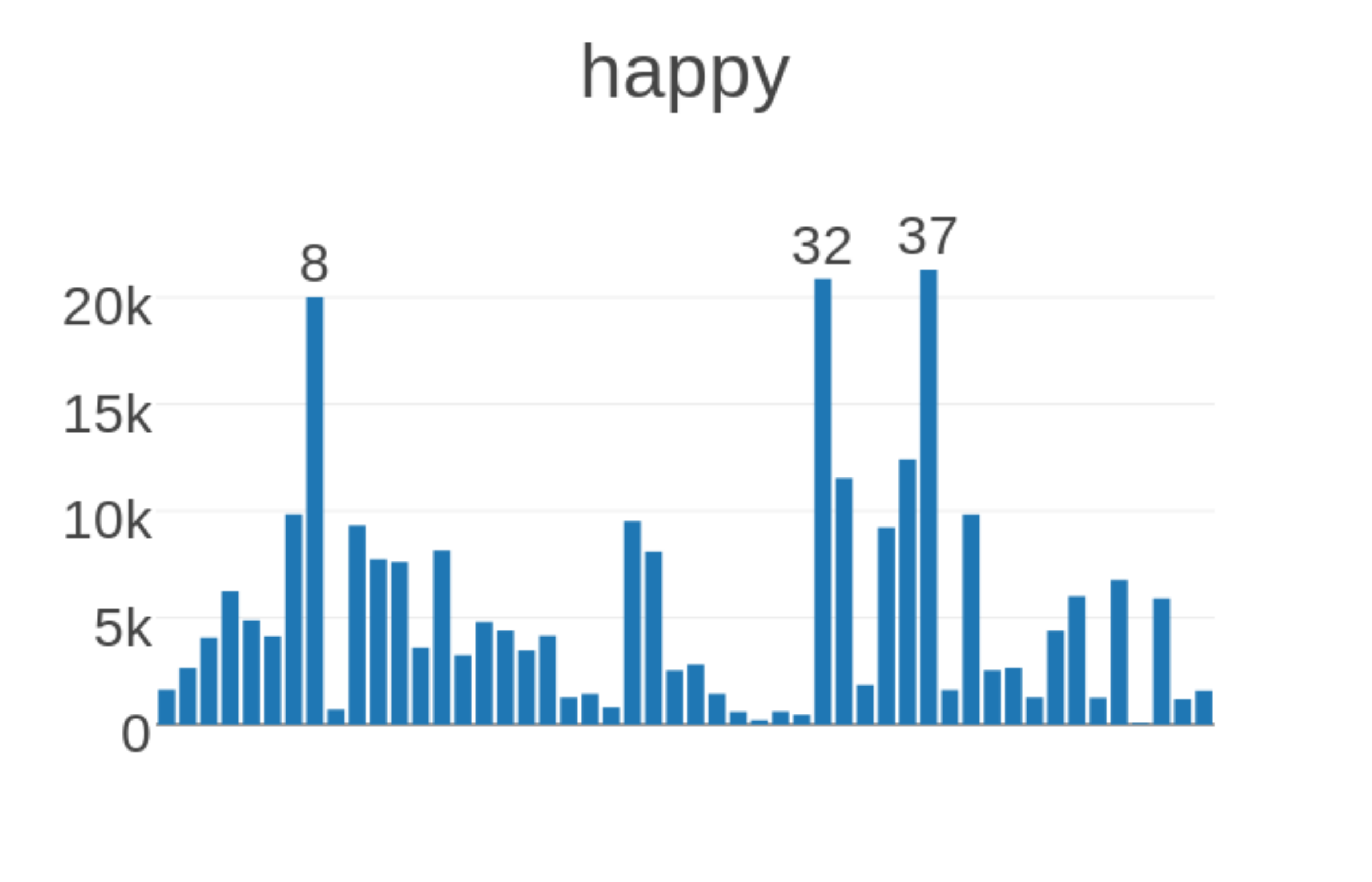} & \includegraphics[width = .23\linewidth]{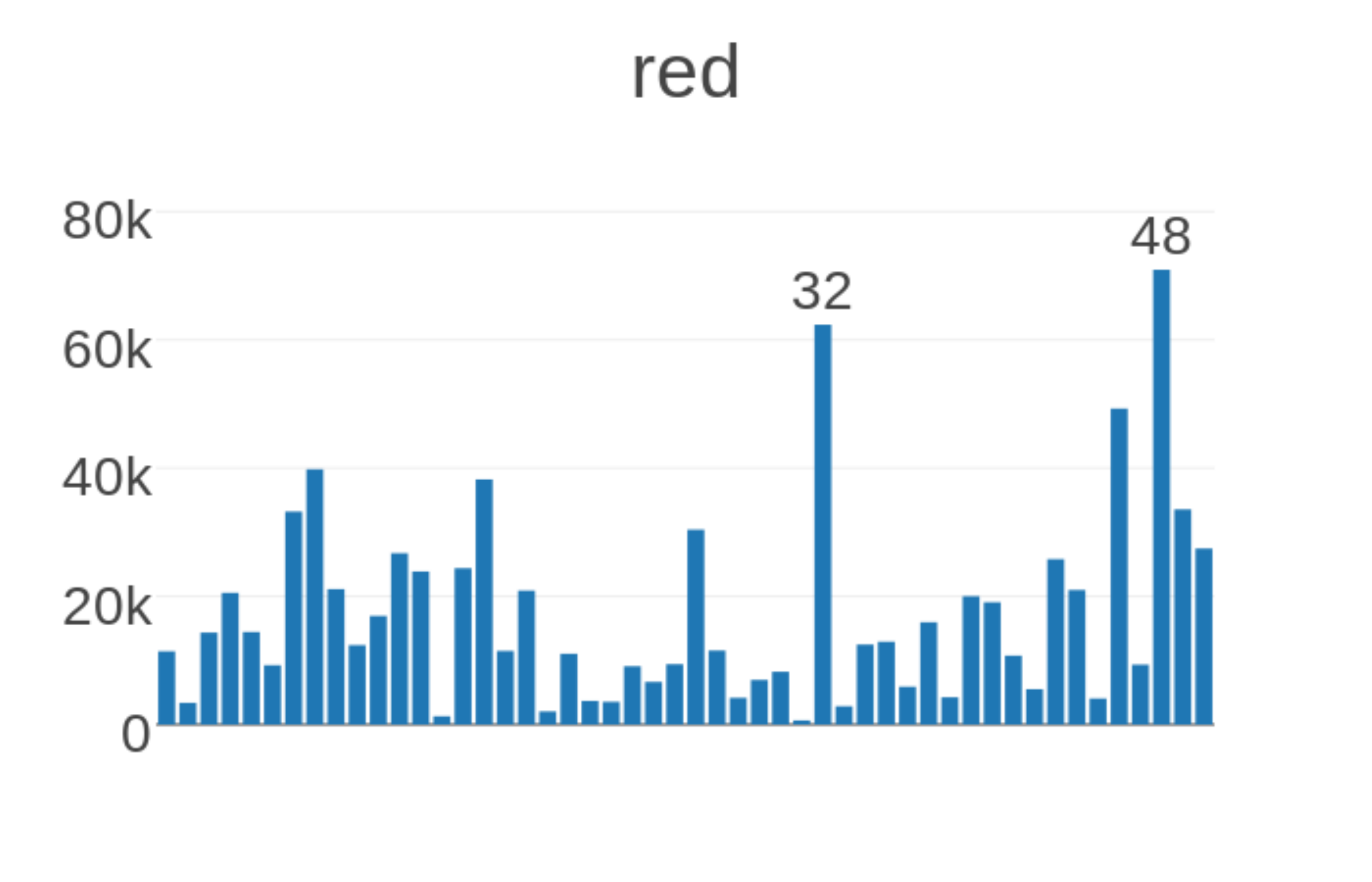} \\

\includegraphics[width = .23\linewidth]{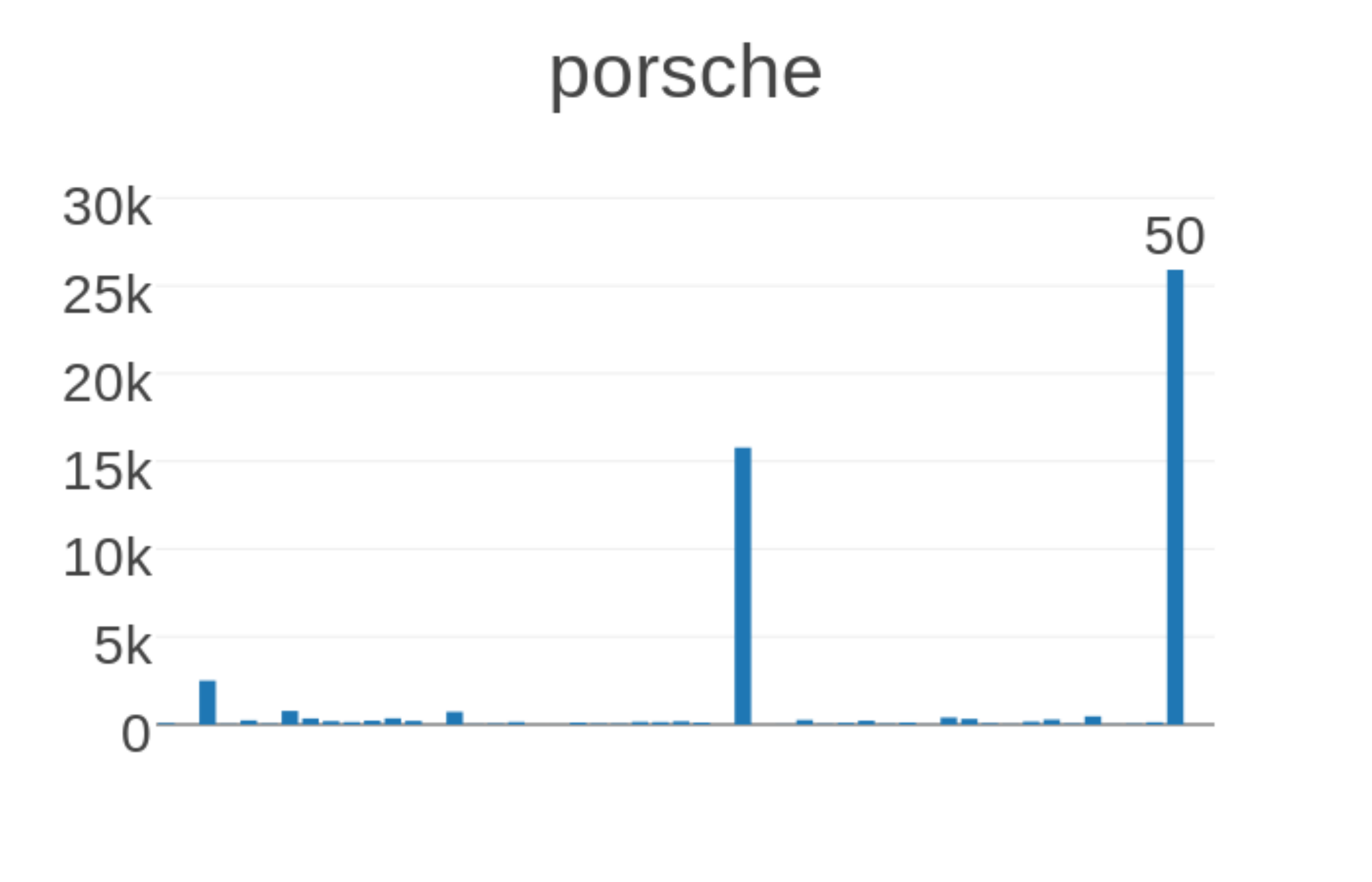} & \includegraphics[width = .23\linewidth]{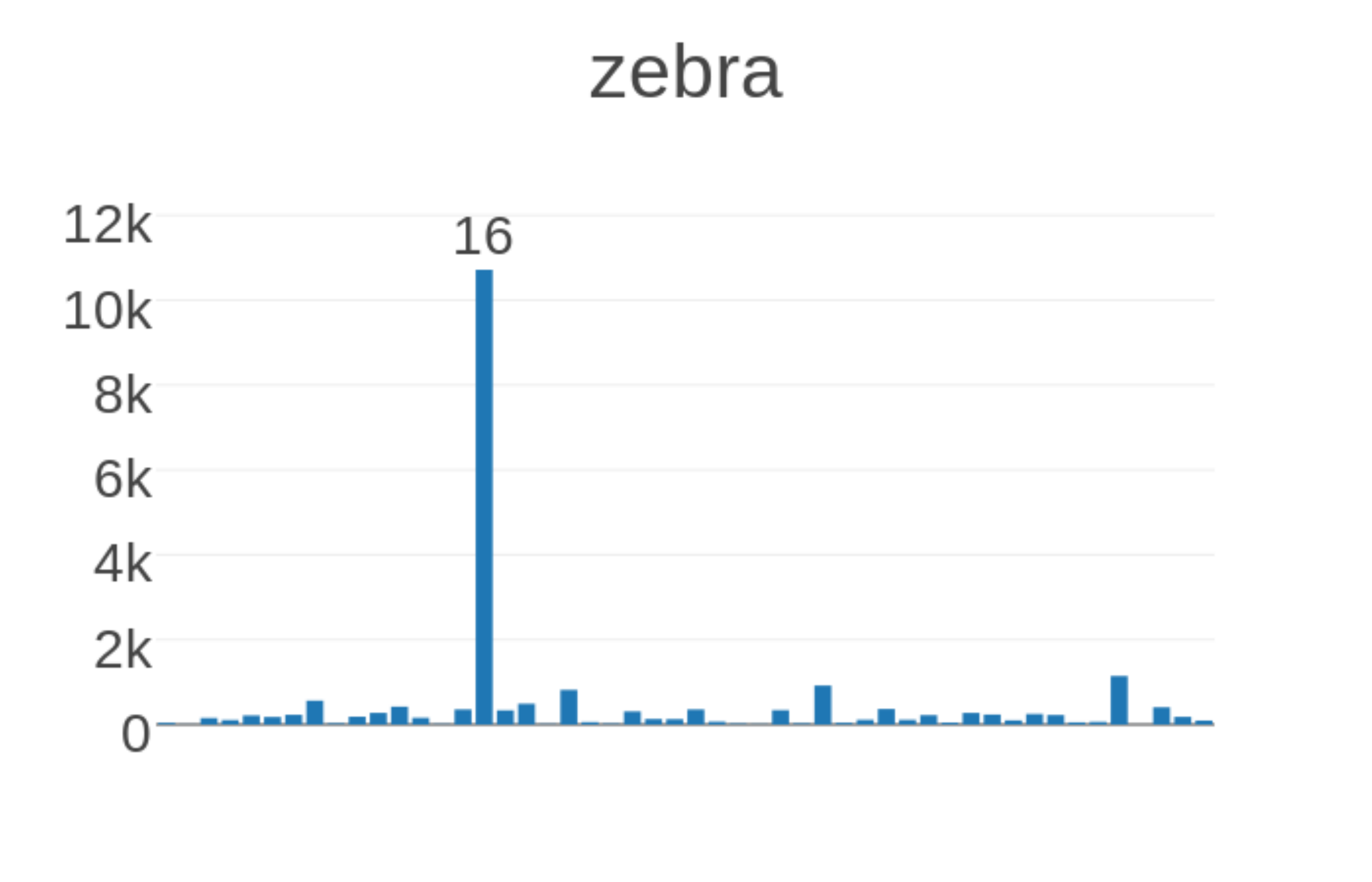} & \includegraphics[width = .23\linewidth]{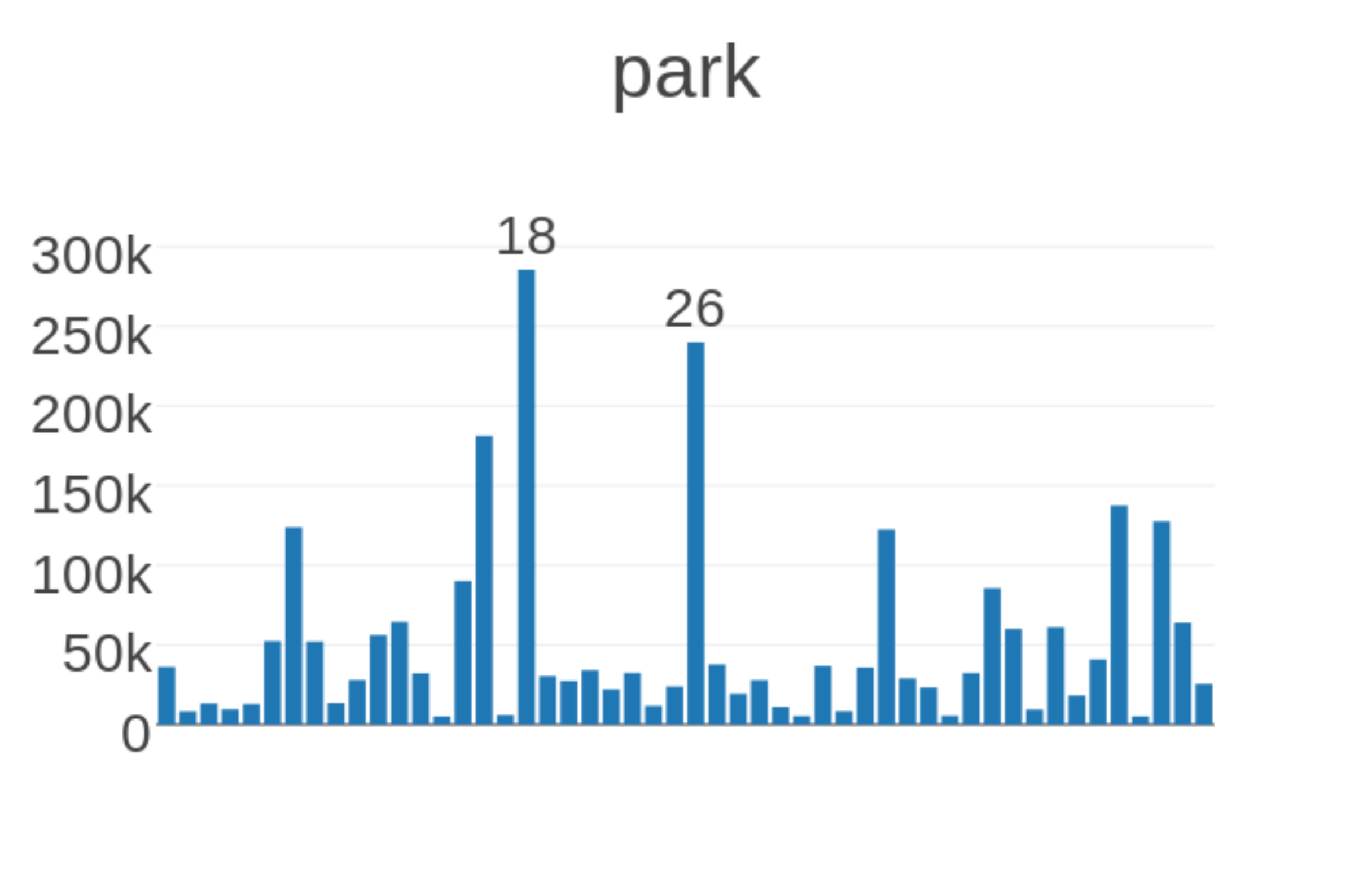} & \includegraphics[width = .23\linewidth]{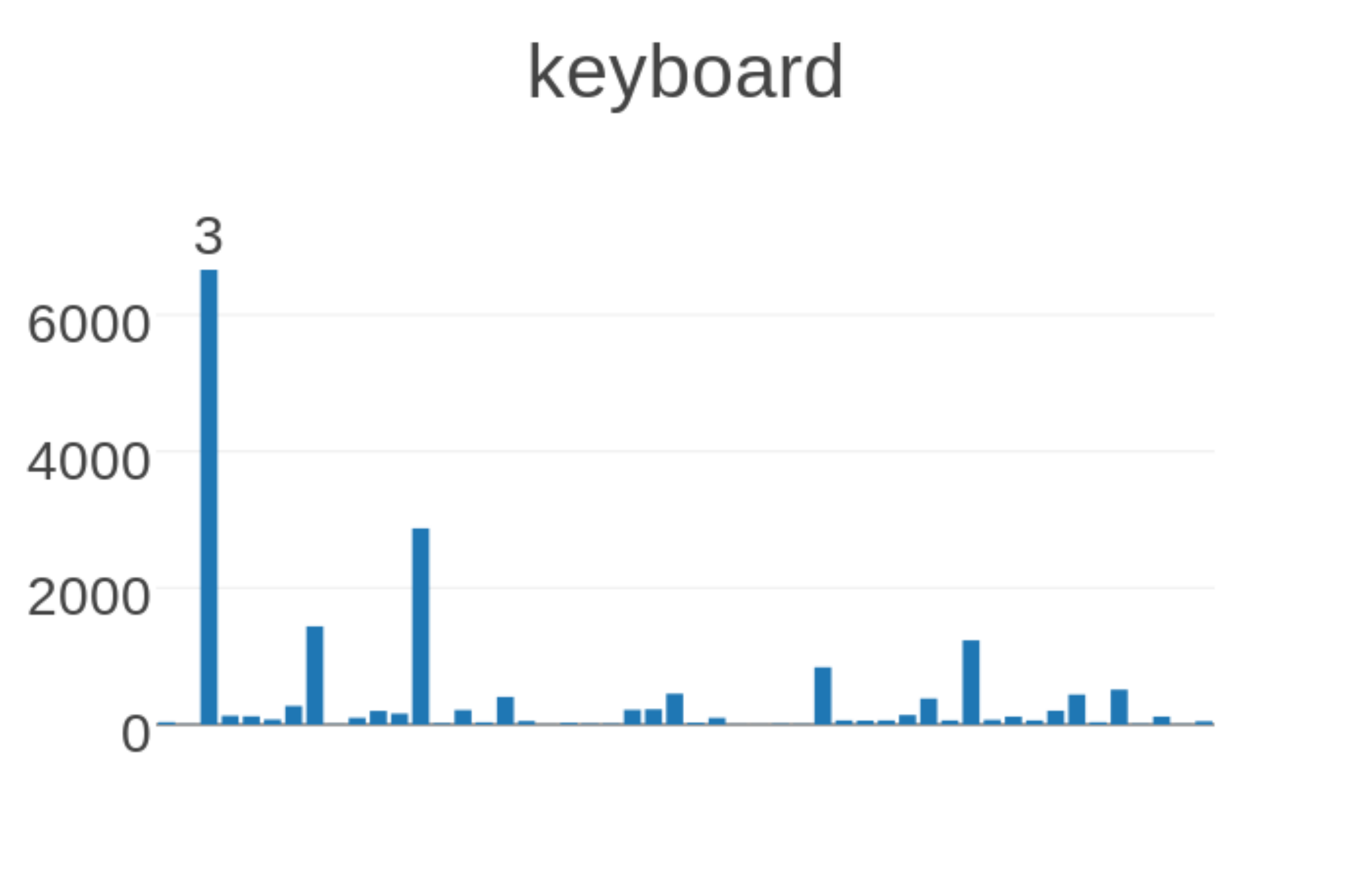} \\
\end{tabular}
\caption{Distribution of tags among clusters.  Each subplot corresponds to a specific word in our dictionary.  Each bar in each histogram corresponds to a specific cluster, and the height of the bar corresponds to the number of times the word appears as a tag in that cluster.   The numbers on top of the bars label the identity of the cluster that bar represents. We can see that some tags are very specific to some clusters, while others are more uniformly spread out.  The 50 clusters are from the feature maps of a ResNet-18 model projected to 256 dimensions}
\label{tab:tag_cluster_hist}
\end{table*}

\subsection{Transfer} 
To test the quality of the features learned by the shared decoder model, we use them as inputs to linear classifiers on the following test datasets: the MIT indoor scenes dataset \cite{indoor}, the SUN scene dataset \cite{SUN}, the CUB birds dataset \cite{CUB}, the Oxford flowers dataset \cite{flowers},
and Stanford 40 action recognition dataset \cite{stanford40}, and Imagenet \cite{imagenet}.  

We use the YFCC100M dataset to train a feature extractor. After training a CNN for the tag prediction task,  we follow standard practice and fix all the layers of the CNN except the decoder.   The output of the layer prior to the decoder taken to be the features for each image in each of the datasets listed above.  We then train a linear classifier (replacing the original model's decoder) for each dataset using these features and the dataset's labels for supervision.

\subsubsection{Transfer Results}
The results of the transfer experiment are shown in Table \ref{tab:transfer}.  There is not a clear pattern of improved performance in tag prediction translating to improved performance in transfer.  Note also that using a single (non-mixture) model {\it trained} on Imagenet does better than any of the models trained on YFCC100M \cite{weakly_supervised_vf}.  Nevertheless, there is a performance gain across datasets for the mixture model, especially for datasets that require fine-grained classification (CUB and Oxford Flowers).   In particular, the results for training the feature extractor on YFCC100M and transferring to Imagenet are encouraging.

\section{Conclusion}

Barring a breakthrough in optimization or a radical change in computer hardware, in order to train convolutional models on datasets with many billions of images without under-fitting, we will need to use specialized model architectures designed for this scale.  In this work we showed that a simple and scalable hard mixture of experts model can significantly raise tag prediction accuracy on large, weakly supervised image collections of 100 million to 500 million images.  In our model, each image is routed to a single expert, and so evaluation has twice the computational cost of the base model;  moreover, it is easy to train the experts in parallel.  We further showed encouraging results on a version of the model where the experts share a decoder, allowing their features to be used for transfer tasks.  

{\small
\bibliographystyle{ieee}
\bibliography{hard_moe_bib}
}

\end{document}